\ifdefined\profilexcludeappendix\else
  \def\includeprofilappendix{1}
\fi

\documentclass{article}

\PassOptionsToPackage{numbers,compress}{natbib}
\ifdefined\profilanonymous
  \usepackage{neurips_2026}
\else
  \usepackage[preprint]{neurips_2026}
\fi

\usepackage[utf8]{inputenc}
\usepackage[T1]{fontenc}
\usepackage[hypertexnames=false,hidelinks]{hyperref}
\usepackage{url}
\usepackage{booktabs}
\usepackage{amsfonts}
\usepackage{amsmath}
\usepackage{amssymb}
\usepackage{nicefrac}
\usepackage{microtype}
\usepackage{xcolor}
\usepackage{graphicx}
\usepackage{multirow}
\usepackage{makecell}
\usepackage{algorithm}
\usepackage{algpseudocode}
\usepackage{enumitem}
\usepackage{tikz}
\usetikzlibrary{arrows.meta,positioning,fit}
\graphicspath{{figures/}}

\ifdefined\profilworkshoptitle
  \workshoptitle{\profilworkshoptitle}
\fi

\title{Drop the Act: Probe-Filtered RL for Faithful Chain-of-Thought Reasoning}

\ifdefined\profilanonymous
  \author{Anonymous Authors}
\else
  \author{Swapnil Parekh \And Naman Goyal}
\fi

\begin{document}

\maketitle

\begin{abstract}
Reasoning models often continue after they have already committed to an answer,
producing \emph{reasoning theater}: deliberative-looking steps that contribute nothing
to correctness. We introduce \textbf{ProFIL} (\textbf{Pro}be-\textbf{Fil}tered
Reinforcement Learning), a drop-in extension to Group Relative Policy Optimization
(GRPO). Prefix counterfactuals identify post-commitment steps; a lightweight probe is
trained once on activations of a base model and then frozen. During RL, the current
policy produces rollouts, while a \emph{separate frozen copy} of the base model
teacher-forces the same rollout text to supply the probe activations. High-theater
rollouts receive zero reward and zero policy advantage.
Across GSM8K, LiveCodeBench, ToolUse, and MMLU-Redux and two model architectures,
ProFIL reduces post-commitment theater by \textbf{11--100\%}, raises faithful fraction
(including $+24$pp on LiveCodeBench under an independent GPT-4.1 judge), and shortens
chains by $4$--$19\%$ in three of four domains, while preserving or improving the
reported task-accuracy metric. A matched length-penalty baseline worsens theater, isolating commitment
detection from generic compression. Frozen-probe audits, an independent pairwise judge,
and inference-time steering controls further support the result and address the
RL-obfuscation concern.
Probe weights, training configurations, evaluation code, and rollout caches are
released for all four domains.
\end{abstract}

\section{Introduction}
\label{sec:intro}

Reinforcement learning has improved reasoning performance
\citep{deepseek2025r1, wei2022chain, kimi2025k15}, but several studies find a gap
between the length of a displayed rationale and the point at which its answer becomes
recoverable.
Biased prompts steer final answers while leaving the chain of thought unchanged
\citep{turpin2023language}; models recover correct answers from truncated chains far earlier
than their length implies \citep{lanham2023measuring}; large-scale audits of production
reasoning models find post-hoc rationalization to be widespread
\citep{anthropic2025faithfulness}.
We study one observable version of this gap: a model can be forced to state the
eventual correct answer from an early prefix, yet continues with redundant derivation or
presentation. We call the resulting suffix \emph{reasoning theater}.

Theater carries a real cost. Post-commitment steps pollute interpretability, introduce
spurious signal into process-reward training \citep{wang2024mathshepherd,
lightman2024prm}, and inflate token cost in agentic deployments where inference budget
matters. Prior work has measured this gap carefully but largely treated it as a
diagnostic. Our goal is to turn the same signal into a post-training objective without
directly rewarding short outputs.

\ifdefined\profilsixpage\else
\begin{figure}[t]
  \centering
  \includegraphics[width=\textwidth]{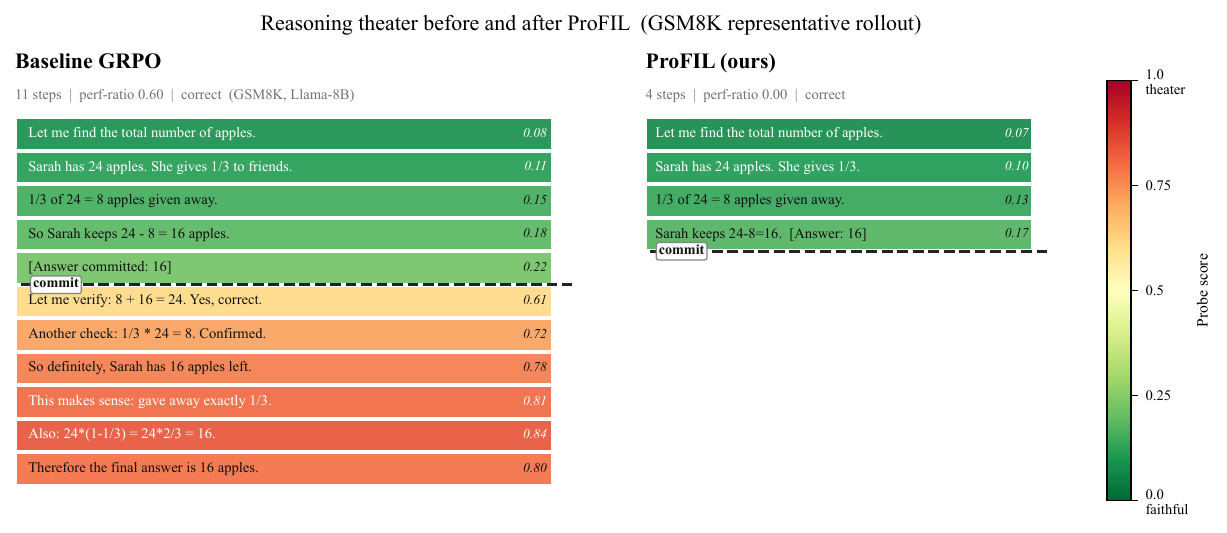}
  \caption{\textbf{ProFIL removes a post-commitment suffix without changing the
    answer.} Rows are steps from a representative GSM8K rollout and numbers are frozen
    probe scores, where larger means more likely post-commitment theater. The dashed line
    is the first prefix with the correct forced answer. Baseline GRPO continues for six
    steps; ProFIL stops there, and both answers are correct.}
  \label{fig:teaser}
\end{figure}
\fi

Faithfulness is not an end-of-rollout verifier reward. ProFIL learns an
internal scorer before RL (Figure~\ref{fig:method}). A trainable policy and a separate
frozen scorer start from the same base weights. The policy samples a rollout; the frozen
copy teacher-forces that exact text, and a frozen probe reads its activations to gate the
GRPO reward. Only the policy changes.

We distinguish the \emph{target construct}, post-commitment steps under prefix
counterfactuals, from the frozen probe score used at scale. To address circularity, we
also use probe-blind and pairwise judges, a frozen-probe audit, and a matched
length-penalty control (\S\ref{sec:exp-corroborate}).

\ifdefined\profilsixpage\else
\textbf{Contributions.}
\begin{itemize}[leftmargin=*]
  \item \textbf{Method and measurement.} ProFIL turns commitment supervision into a
    frozen reward filter, keeps policy and scorer parameters separate, and distinguishes
    temporal from logical faithfulness.
  \item \textbf{Evidence.} Theater falls 11--100\% across four domains and two
    architectures. Independent judges, probe audits, and controls support the result;
    accuracy is reported under an explicit per-domain protocol.
  \item \textbf{Mechanism and artifacts.} ProFIL collapses a bimodal theater mode with
    a visible linguistic signature. We release probes, configurations, code, and caches.
\end{itemize}
\fi

\section{Related Work}
\label{sec:related}

\ifdefined\profilsixpage
Faithfulness studies expose gaps between displayed reasoning and answers
\citep{turpin2023language,lanham2023measuring,anthropic2025faithfulness}; our temporal
question complements logical faithfulness \citep{lyu2023faithful} and turns the
diagnostic into a training signal. Concision methods reward length directly
\citep{dai2025sgrpo,dumitru2025conciserl,zhang2025grpolead,kimi2025k15,aggarwal2025l1},
whereas ProFIL targets commitment. Probes recover and steer latent properties
\citep{burns2023ccs,azaria2023internal,li2023iti,rimsky2024caa}; unlike RLFR
\citep{goodfire2026rlfr}, ProFIL scores a frozen base rather than the changing policy.
Process reward models score step correctness \citep{wang2024mathshepherd,lightman2024prm};
our complementary scorer asks whether the step occurs after commitment.
\else
\textbf{Chain-of-thought faithfulness.}
Biased-prompt and forced-answer studies expose gaps between displayed reasoning and
answers \citep{turpin2023language,lanham2023measuring}; production audits find the gap
widespread \citep{anthropic2025faithfulness}. We act on this diagnostic during training.
Our temporal question, whether writing continues after commitment, complements logical
faithfulness, whether the stated reasoning entails the answer \citep{lyu2023faithful}.

\textbf{Concise and early-terminating reasoning.}
S-GRPO \citep{dai2025sgrpo}, ConciseRL \citep{dumitru2025conciserl}, and GRPO-LEAD
\citep{zhang2025grpolead}, brevity penalties \citep{kimi2025k15}, and L1
\citep{aggarwal2025l1} reward length or concision. Our matched ablation instead becomes
\emph{more} performative on LiveCodeBench (\S\ref{sec:exp-robust}). ProFIL targets
commitment; shortening is a byproduct.

\textbf{Probes as reward signals.}
Probes recover latent knowledge \citep{burns2023ccs,azaria2023internal} and steer
inference \citep{li2023iti,rimsky2024caa}. RLFR brought multi-head attention probes into
RL \citep{goodfire2026rlfr}; ProFIL instead reads a \emph{frozen base}, not the changing
policy. Unlike CAA or ITI, which leave weights untouched, ProFIL changes what the policy
learns.

\textbf{Process reward models.}
Process Reward Models (PRMs) score step correctness
\citep{wang2024mathshepherd,lightman2024prm}; ProFIL scores commitment. A step can be
correct yet post-commitment, so PRMs and ProFIL are complementary.
\fi

\section{Background}
\label{sec:bg}

\textbf{GRPO.}
GRPO \citep{shao2024deepseekmath, deepseek2025r1, kimi2025k15} samples $K$ rollouts per
prompt, scores each with a \emph{verifier} (a binary correctness checker, e.g., string
match for GSM8K, test-case execution for LiveCodeBench), and updates the policy via
group-relative advantage. Because the verifier cannot score non-verifiable behaviors
like faithfulness, ProFIL gates which rollouts contribute to the update
(\S\ref{sec:method}). We distinguish the forced-answer target, the probe-based
evaluation metric, and the training threshold formally in \S\ref{sec:method}.

\section{ProFIL: Probe-Filtered Reinforcement Learning}
\label{sec:method}

\begin{figure}[!htbp]
  \centering
  \begin{tikzpicture}[
      x=1cm,y=1cm,
      every node/.style={font=\small},
      arr/.style={-{Latex[length=2.7mm,width=2.0mm]},line width=0.9pt,draw=black},
      frozen/.style={draw=blue!60!black,double=blue!60!black,double distance=0.45pt,
                    fill=blue!4,rounded corners=1.2pt,align=center,inner sep=3.8pt,minimum height=10mm},
      train/.style={draw=orange!75!black,line width=1.1pt,fill=orange!5,
                   rounded corners=1.2pt,align=center,inner sep=3.8pt,minimum height=10mm},
      operation/.style={draw=black!55,fill=white,rounded corners=1.2pt,align=center,inner sep=3.5pt,
                       minimum height=10mm}]

    \node[anchor=west,font=\normalsize\bfseries] at (0,5.35)
      {(a) Offline: build one fixed scorer};

    \node[anchor=north west,inner sep=0pt] (labels) at (0,4.98) {%
      \setlength{\tabcolsep}{4pt}%
      \renewcommand{\arraystretch}{1.16}%
      \begin{tabular}{lccccc}
        \toprule
        Prefix & $y_1$ & $y_2$ & \textbf{$y_3$} & $y_4$ & $y_5$ \\
        Forced answer correct? & no & no & \textbf{yes} & yes & yes \\
        Probe target $\rho_t$ & 0 & 0 & 0 & \textbf{1} & \textbf{1} \\
        \bottomrule
      \end{tabular}};
    \node[anchor=west,font=\footnotesize] at (0,3.08)
      {Commitment $c=3$; targets are 1 only for $t>c$.};

    \node[frozen,minimum width=2.75cm] (baseoff) at (7.65,3.17)
      {base $\pi_0$ \textbf{(frozen)}\\read activations $h_t$};
    \node[train,minimum width=3.05cm] (fitprobe) at (11.05,4.10)
      {probe $f_\phi$\\train once, then \textbf{freeze}};
    \draw[arr] (labels.east) -- (fitprobe.west);
    \draw[arr] (baseoff.east) -- (fitprobe.south west);

    \draw[black!35,line width=0.6pt] (0,2.62)--(12.85,2.62);
    \node[anchor=west,font=\normalsize\bfseries] at (0,2.30)
      {(b) Online: score each rollout inside GRPO};

    \node[train,minimum width=2.65cm] (policy) at (1.30,1.35)
      {\textbf{trainable} policy $\pi_\theta$\\sample $K$ rollouts};
    \node[frozen,minimum width=2.90cm] (base) at (4.65,1.35)
      {\textbf{frozen} base $\pi_0$\\teacher-force rollout};
    \node[frozen,minimum width=2.55cm] (probeonline) at (7.95,1.35)
      {\textbf{frozen} probe $f_\phi$\\score $\bar p^{(k)}$};
    \node[operation,text width=3.40cm] (gate) at (7.95,0.02)
      {reward gate\\keep iff $\bar p^{(k)}<\tau$; else reward 0};
    \node[train,minimum width=2.05cm] (update) at (11.60,0.02)
      {GRPO\\update $\theta$};
    \draw[arr] (policy.east)--(base.west);
    \draw[arr] (base.east)--(probeonline.west);
    \draw[arr] (probeonline.south)--(gate.north);
    \draw[arr] (gate.east)--(update.west);
    \draw[arr] (update.south) -- (11.60,-0.78) -- (1.30,-0.78) -- (policy.south);
    \node[font=\footnotesize,fill=white,inner sep=1pt] at (6.10,-0.78)
      {only $\theta$ receives gradients};
  \end{tikzpicture}
  \caption{\textbf{The scorer is learned once, then held fixed.} (a) Forced answering
    finds the first prefix that already yields the correct answer. Later steps receive
    target 1. These labels train a probe on frozen-base activations. (b) During GRPO,
    the policy produces a rollout, while a separate frozen base and the frozen probe
    score that same text. The filter keeps or zeros the verifier reward. Only the
    policy parameters $\theta$ receive gradients.}
  \label{fig:method}
\end{figure}

\subsection{Target construct: commitment under forced answering}

A reasoning policy $\pi_\theta$ produces a chain
$y=(y_1,\ldots,y_T)$ and answer $a$ for input $x$. We divide the chain at natural step
boundaries. Let $F(x,y_{1:t})$ append a stop-and-answer instruction to prefix
$y_{1:t}$ and greedily decode an answer. With reference answer $a^\star(x)$, the
\emph{commitment point} is
\[
 c(y;x)=\min\{t:F(x,y_{1:t})=a^\star(x)\}.
\]
Every step after $c$ is performative because the correct answer was already
extractable. For a committed chain, the forced-answer target ratio is
\[
 \operatorname{perf}_{\rm FA}(y;x)=\frac{T-c(y;x)}{\max(T-1,1)}.
\]
If no prefix produces the correct answer, the chain is \emph{uncommitted} and receives
target ratio zero by convention. This does not mean the chain is faithful; it means
there is no forced-answer commitment point from which to define a suffix.

This definition measures \emph{temporal faithfulness}: does the model keep writing
after its answer is recoverable? Logical faithfulness asks whether the written premises
entail the answer. Causal faithfulness asks whether the visible steps caused the
model's internal computation. These are different questions. A valid proof can still
contain a long post-commitment restatement, while a short invalid proof can be
temporally faithful.

\textbf{How the labels are made.} We run forced answering at every step boundary. For
GSM8K and MMLU-Redux, the verifier checks the normalized number or answer choice. For
LiveCodeBench, it executes the forced code against test cases. For ToolUse, it matches
the tool name and arguments. ToolUse first looks for a JSON tool-call closure followed
by a terminal marker and falls back to forced-answer matching when no such closure is
present. The first correct prefix fixes $c$; every later step gets label 1 and every
earlier step gets label 0. No human annotation is needed. The policy never sees the
forced-answer prompt during RL.

\subsection{Frozen-base attention probe}

We train an RLFR-style gated multi-head attention probe \citep{goodfire2026rlfr} on
the labeled base-model rollouts. Its keys and values come from LayerNorm-projected
residual-stream activations at two layers of the frozen base. Eight learned queries
attend over each reasoning step, sigmoid gates combine the heads, and a scalar output
gives the step's probability of being performative. Binary cross-entropy trains the
probe once. We then freeze it. Held-out area under the receiver operating
characteristic curve (AUROC) is
summarized below rather than compressed into a slash-separated list.

\begin{center}
  \setlength{\tabcolsep}{7pt}
  \small
  \begin{tabular}{l c c c c}
    \toprule
    Domain & GSM8K & LiveCodeBench & ToolUse & MMLU-Redux \\
    \midrule
    Held-out AUROC & 0.924 & 0.928 & 0.976 & 0.991 \\
    \bottomrule
  \end{tabular}
\end{center}

During RL, the current policy generates a rollout. We then teacher-force that same
text through a separate frozen copy of the base model. The probe reads this frozen
copy's activations and produces per-step probabilities $p_{1:T}$. Their mean
$\bar p$ is the rollout score. The frozen base does not generate a second answer and
the probe is not a separate language model. It is a small classifier over internal
activations. Keeping both the base and probe fixed gives the policy a stable scoring
boundary throughout training.

\textbf{Operational evaluation metric.}
The reported \emph{performativity ratio} (PR) is probe-based, not
$\operatorname{perf}_{\rm FA}$. With evaluation cutoff $\eta=0.9$, we threshold the
frozen probe score for each evaluation unit and compute the fraction of positive units
in a chain; PR is its mean over chains. The \emph{operational faithful fraction} is the
share of chains with no probe-positive unit. GSM8K, LiveCodeBench, and MMLU-Redux use
one pooled chain score, so PR is a binary rate and receives a Wilson interval. ToolUse
retains multiple planning blocks, reports the mean within-chain positive fraction, and
uses bootstrap intervals. Here $\eta$ defines the reported metric, while the separate
$\tau$ below gates training. A sweep over $\eta$ is reported in
\S\ref{sec:exp-robust}.

\subsection{The probe filter}

GRPO \citep{shao2024deepseekmath} samples $K$ rollouts per prompt and obtains verifier
reward $r^{(k)}$. ProFIL keeps rollout $k$ when its mean predicted performativity is
below $\tau$. The filter appears directly in both the reward and advantage:
\[
 \widetilde r^{(k)}=\mathbf{1}[\bar p^{(k)}<\tau]r^{(k)},\qquad
 \widehat A^{(k)}=\mathbf{1}[\bar p^{(k)}<\tau]
 \frac{\widetilde r^{(k)}-\mu(\widetilde r)}
 {\sigma(\widetilde r)+\epsilon}.
\]
A high-theater correct rollout contributes zero reward to group normalization and
zero advantage to the policy update. Incorrect rollouts already receive zero verifier
reward. The threshold $\tau$ is the only new hyperparameter. Because a rollout is kept
when $\bar p<\tau$, a smaller $\tau$ is more aggressive. LiveCodeBench uses $0.2$;
the other domains use $0.5$. No gradient enters the frozen base or probe.

The training signal is entirely this reward-and-advantage filter. The policy never sees
the forced-answer prompt, the probe score, or an instruction to be concise.

\begin{table}[!t]
  \centering
  \small
  \caption{\textbf{ProFIL reduces theater by 11--100\% across all four domains while
    preserving or improving the reported accuracy metric.}
    PR is the frozen-probe positive rate at evaluation cutoff $\eta=0.9$; operational
    faithful fraction is the share of chains with no positive unit. Accuracy is
    greedy@1 except LiveCodeBench, where the table reports
    the sampled@1 in-training peak; \S\ref{sec:lcb-case} reports both protocols.
    Bracketed ranges are 95\% confidence intervals (CIs). Binary rates and task accuracy use Wilson CIs; ToolUse PR
    uses chain-level pooling and a bootstrap CI.}
  \label{tab:main}
  \resizebox{\textwidth}{!}{%
  \begin{tabular}{l l c c c c c}
    \toprule
    Domain & Condition &
      \makecell{PR $\downarrow$\\{\scriptsize [95\% CI]}} &
      \makecell{Relative\\change} &
      \makecell{Faithful\\frac. $\uparrow$} &
      \makecell{Task accuracy $\uparrow$\\{\scriptsize [95\% CI]}} &
      \makecell{Think\\chars} \\
    \midrule
    GSM8K
      & Baseline & 0.006 {\scriptsize[0.002--0.017]} & -- & 0.994 & 0.778 {\scriptsize[0.740--0.812]} & 2{,}351 \\
      & \textbf{ProFIL} & \textbf{0.000 {\scriptsize[0.000--0.008]}} & \textbf{${\sim}100\%$} & \textbf{1.000} & \textbf{0.830 {\scriptsize[0.795--0.860]}} & \textbf{1{,}895} \\
    \midrule
    \makecell[l]{Live-\\CodeBench$^\dagger$}
      & Baseline & 0.267 {\scriptsize[0.199--0.349]} & -- & 0.733 & 0.30 (peak) & 12{,}382 \\
      & \textbf{ProFIL} & \textbf{0.076 {\scriptsize[0.042--0.135]}} & \textbf{$-72\%$} & \textbf{0.924} & \textbf{0.37 (peak)} & \textbf{11{,}875} \\
    \midrule
    ToolUse$^\ddagger$
      & Baseline & 0.527 {\scriptsize[0.48--0.57]} & -- & 0.029 & 0.529 {\scriptsize[0.41--0.64]} & 1{,}248 \\
      & \textbf{ProFIL} & \textbf{0.468 {\scriptsize[0.41--0.53]}} & \textbf{$-11\%$} & \textbf{0.147} & 0.559 {\scriptsize[0.44--0.67]} & \textbf{1{,}018} \\
    \midrule
    MMLU-Redux
      & Baseline & 0.016 {\scriptsize[0.008--0.031]} & -- & 0.984 & 0.624 {\scriptsize[0.581--0.665]} & 4{,}973 \\
      & \textbf{ProFIL} & \textbf{0.008 {\scriptsize[0.003--0.020]}} & \textbf{$-50\%$} & \textbf{0.992} & 0.624 {\scriptsize[0.581--0.665]} & 5{,}240 \\
    \bottomrule
  \end{tabular}}%
  \\
  \footnotesize
  $^\dagger$Performativity uses greedy@1; accuracy is sampled@1 in-training peak.
  $^\ddagger$ToolUse uses chain-level pooling.
\end{table}

\ifdefined\profilsixpage\else
\begin{figure}[!t]
  \centering
  \includegraphics[width=\textwidth]{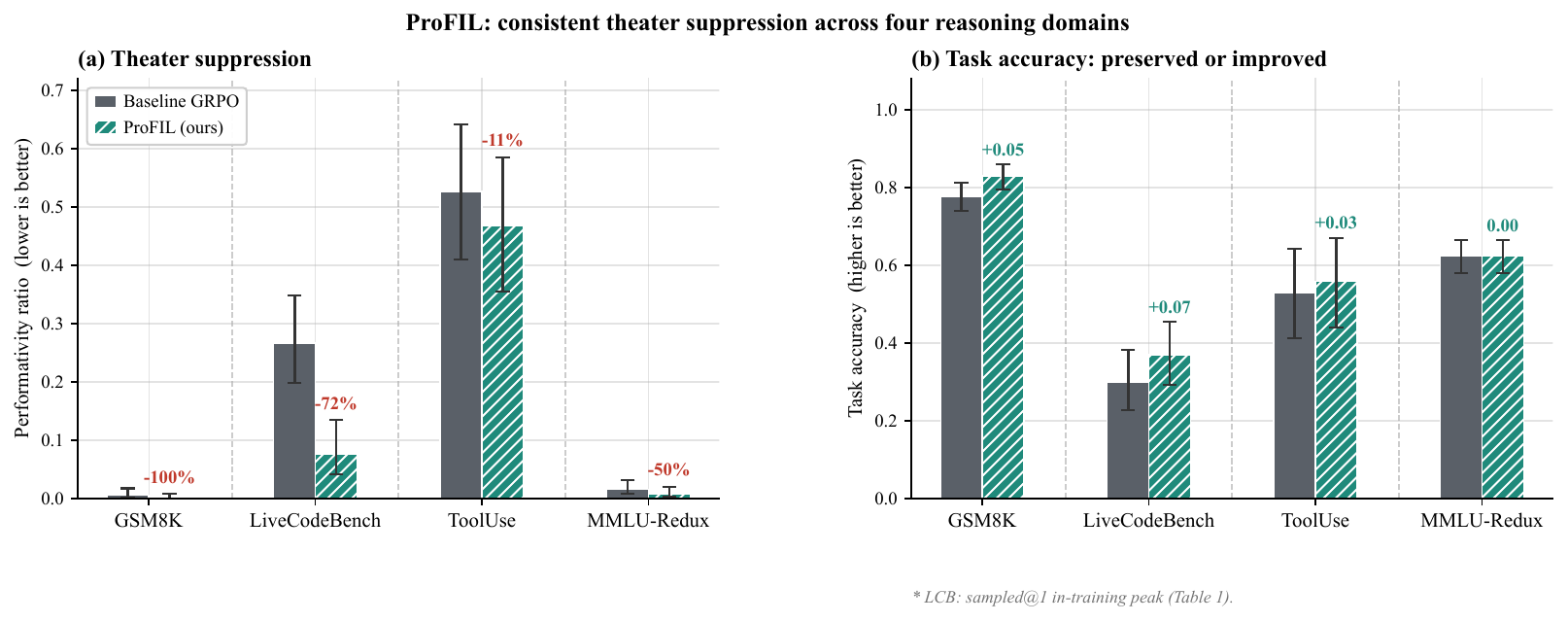}
  \caption{\textbf{ProFIL suppresses theater while preserving or improving the
    reported task metric in all four domains.} Left: frozen-probe positive rate
    falls by 11--100\% (lower is better). Right: the reported task metric is
    preserved or improved (higher is better). Error bars are 95\% CIs; exact
    estimates and protocols appear in Table~\ref{tab:main}.}
  \label{fig:main-results}
\end{figure}
\fi

\section{Experiments}
\label{sec:exp}

\subsection{Setup}
\label{sec:exp-setup}

\textbf{Models.} DeepSeek-R1-Distill-Llama-8B for GSM8K and LiveCodeBench;
DeepSeek-R1-Distill-Qwen-7B for ToolUse and MMLU-Redux. This pairing follows existing
per-backbone training infrastructure and checkpoint availability. Each matched
baseline uses the same backbone, data, optimizer, context limit, and $K=8$ rollouts per
prompt. ProFIL adds only the frozen scorer and threshold $\tau$. We use $\tau=0.2$ on
sparse-reward LiveCodeBench and $0.5$ elsewhere. Training runs for three epochs except
MMLU-Redux, which runs for 256 GRPO steps. All runs use AdamW with a cosine schedule;
the released configurations contain exact learning rates, batch sizes, and optimizer
settings.

\textbf{Datasets.}
\ifdefined\profilsixpage
We use GSM8K (7,440 train/500 held out; boxed-answer match), LiveCodeBench v6
(920/all 131; execution), ToolUse (4,046/68; exact tool and argument match), and
MMLU-Redux (6,144/500 across 57 subjects; answer-choice match)
\citep{cobbe2021gsm8k,jain2024livecodebench,gema2025mmluredux,hendrycks2021mmlu}.
\else
\begin{itemize}[leftmargin=*, itemsep=1pt, topsep=2pt]
  \item \textbf{GSM8K} \citep{cobbe2021gsm8k}: grade-school math, with 7,440 training
    and 500 held-out problems; boxed-answer match.
  \item \textbf{LiveCodeBench v6} \citep{jain2024livecodebench}: competitive
    programming, with 920 training and all 131 held-out problems; test-case execution.
  \item \textbf{ToolUse}: 4,046 training and 68 held-out multi-step agent traces;
    exact tool name and argument match.
  \item \textbf{MMLU-Redux} \citep{gema2025mmluredux, hendrycks2021mmlu}: 500
    held-out questions across 57 subjects; answer-choice match.
\end{itemize}
\fi
Evaluation uses vLLM \citep{kwon2023vllm}. All metrics use greedy@1 except the
explicitly marked LiveCodeBench sampled@1 training peak. We report 95\% Wilson
intervals \citep{wilson1927ci} for binary rates and bootstrap intervals for ToolUse.

\ifdefined\profilsixpage
The frozen scorer reads residual-stream layers 13 and 19 for Llama-8B and layers 14
and 21 for Qwen-7B. Context limits are 8,192 tokens on LiveCodeBench and 4,096
elsewhere; the released configurations specify all remaining optimizer settings.
\else
\begin{table}[t]
  \centering
  \small
  \caption{\textbf{Training and scoring configuration.} All conditions use $K=8$
    rollouts per prompt. The matched baseline shares every setting except the probe
    filter. ``Layers'' are the two frozen-base residual-stream layers read by the probe.}
  \label{tab:config}
  \resizebox{\textwidth}{!}{%
  \begin{tabular}{l l r r c c c c}
    \toprule
    Domain & Backbone & Train & Eval & Training & Context & Layers & $\tau$ \\
    \midrule
    GSM8K & Llama-8B & 7,440 & 500 & 3 epochs & 4,096 & 13, 19 & 0.5 \\
    LiveCodeBench & Llama-8B & 920 & 131 & 3 epochs & 8,192 & 13, 19 & 0.2 \\
    ToolUse & Qwen-7B & 4,046 & 68 & 3 epochs & 4,096 & 14, 21 & 0.5 \\
    MMLU-Redux & Qwen-7B & 6,144 & 500 & 256 steps & 4,096 & 14, 21 & 0.5 \\
    \bottomrule
  \end{tabular}}
\end{table}
\fi

\textbf{Forced-answer details.}
At each step boundary we append: ``Stop reasoning now. Based on what you have written
above, give your final answer in the format \texttt{\textbackslash boxed\{\}}.'' We greedily
decode at most 64 new tokens. LiveCodeBench instead executes the forced code and treats
pass rate above $0.5$ as correct. These calls build probe labels offline and are never
shown to the policy during RL.

\textbf{Training cost.}
Filtering adds one frozen-model forward pass and no backward pass or optimizer state.
On an A100-40GB, scoring one LiveCodeBench rollout takes about 319\,ms, compared with
about 533\,ms for a comparable policy forward-backward update. The eight RL runs used
about 480 H100-80GB GPU-hours in total. Probe training used about 2 GPU-hours per domain,
evaluation about 8 per domain, and the reported ablations about 200 additional GPU-hours.

\subsection{ProFIL Suppresses Theater on All Four Domains}
\label{sec:exp-main}

\begin{table}[t]
  \centering
  \small
  \caption{\textbf{Penalizing length makes theater worse on LiveCodeBench.}
    The matched length penalty produces shorter chains but raises performativity.
    ProFIL lowers performativity without directly rewarding brevity. Bracketed
    ranges are 95\% confidence intervals.}
  \label{tab:lp}
  \begin{tabular}{l c c c c}
    \toprule
    Condition &
      \makecell{Perf. ratio $\downarrow$\\\scriptsize [95\% CI]} &
      \makecell{Faithful\\frac.\ $\uparrow$} &
      \makecell{Accuracy\\(greedy@1) $\uparrow$} &
      \makecell{Think\\chars} \\
    \midrule
    Baseline       & 0.267 {\scriptsize[0.199--0.349]} & 0.733 & 0.214 & 12{,}382 \\
    Len.\ Penalty  & 0.374 {\scriptsize[0.296--0.459]} & 0.626 & 0.229 & \textbf{11{,}349} \\
    \textbf{ProFIL} & \textbf{0.076 {\scriptsize[0.042--0.135]}} & \textbf{0.924} & 0.183 & 11{,}875 \\
    \bottomrule
  \end{tabular}
\end{table}

\textbf{Theater suppression}
\ifdefined\profilsixpage
(Table~\ref{tab:main}).
\else
(Table~\ref{tab:main}; Figure~\ref{fig:main-results}).
\fi
Performativity drops in every domain: GSM8K $0.006 \rightarrow 0.000$
(${\sim}100\%$), LiveCodeBench $0.267 \rightarrow 0.076$ ($-72\%$,
CIs non-overlapping), ToolUse $0.527 \rightarrow 0.468$ ($-11\%$ chain-level), and
MMLU-Redux $0.016 \rightarrow 0.008$ ($-50\%$). Faithful-fraction rises in every domain
as well, with the largest jump on ToolUse ($0.029 \rightarrow 0.147$, a $5\times$ gain)
and on LiveCodeBench ($0.733 \rightarrow 0.924$).
GSM8K accuracy rises by 5.2 points, ToolUse by 3 points, and MMLU-Redux stays at exact
parity. Reasoning becomes 19\% shorter on GSM8K, 4\% shorter on LiveCodeBench, and 18\%
shorter on ToolUse. MMLU-Redux is the exception: it halves theater at equal accuracy
while mean length rises by 5\%.

\label{sec:lcb-case}
\textbf{LiveCodeBench protocol.}
The training-time sampled validation peak rises from 0.30 (baseline step 320) to 0.37
(ProFIL step 440) and is the accuracy metric in Table~\ref{tab:main}. Validation was
logged every 20 steps with four sampled rollouts per problem. This peak is descriptive
training telemetry, not a final greedy held-out estimate. At the final checkpoint,
greedy@1 on all 131 test problems is 0.214 for the baseline,
0.229 for length penalty, and 0.183 for ProFIL. The theater result does not depend on
which accuracy protocol is chosen, but the accuracy conclusion does. We claim
an accuracy gain for the reported sampled-peak protocol, not for every checkpoint and
decoding rule.

\ifdefined\profilsixpage\else
\begin{figure}[t]
  \centering
  \includegraphics[width=\textwidth]{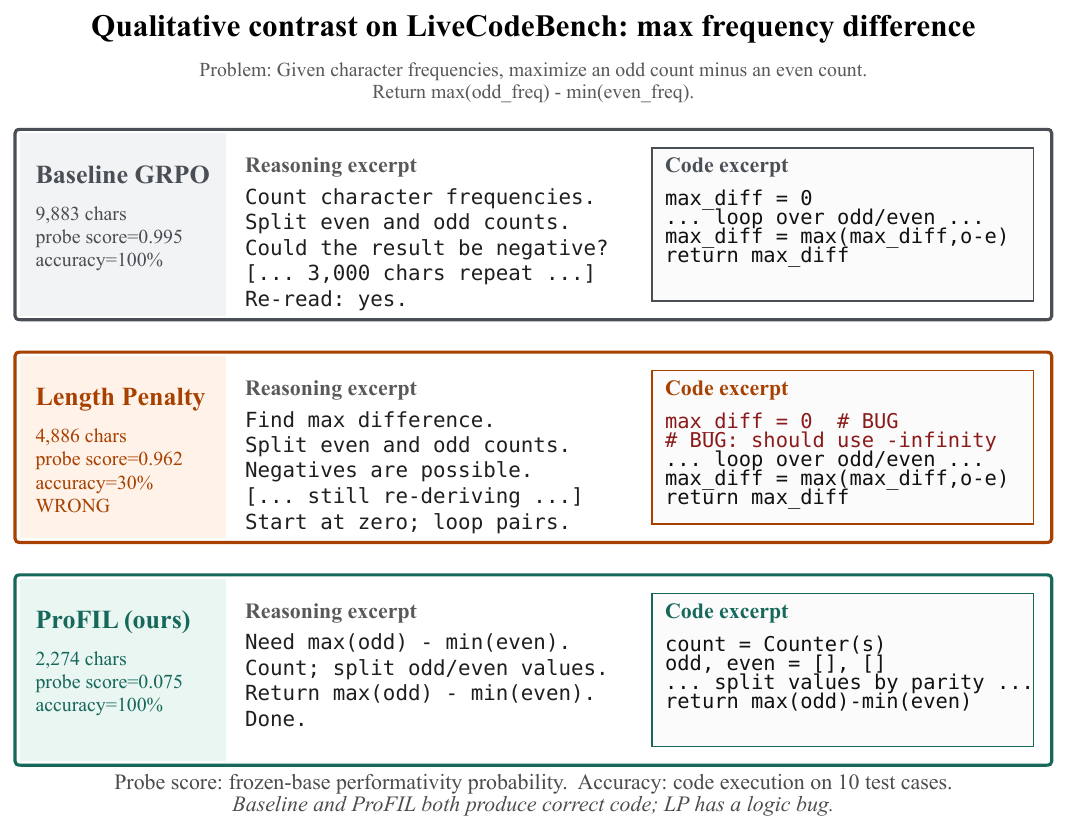}
  \caption{\textbf{A concrete LiveCodeBench comparison.} We show the highest-contrast
    example among 14 high-theater baseline cases. The baseline repeatedly revisits a
    solved edge case (9,883 characters, probe score 0.995). Length penalty is shorter
    but remains theatrical and introduces an initialization bug (4,886 characters,
    score 0.962). ProFIL reaches the correct five-line program directly (2,274
    characters, score 0.075).
    This example is illustrative rather than a population estimate.}
  \label{fig:qualitative}
\end{figure}
\fi

\subsubsection{ToolUse: Chain-Level Theater and a Fivefold Faithful-Fraction Gain}
\label{sec:tooluse}

On ToolUse, theater appears as extra planning blocks after the model has already
committed to a complete tool call. ProFIL raises faithful fraction fivefold, from
$0.029$ to $0.147$.

\textbf{ProFIL increases single-tool-call resolutions $\boldsymbol{8\times}$.}
Single-thinking-block resolutions rise from $1/68$ to $8/68$. In these rollouts the
model identifies the right tool and acts without generating a second block to reconsider
the same plan. Genuine multi-step plans remain possible because the filter targets
post-commitment behavior rather than the number of tool calls.

\subsection{Length Compression Is Not the Operative Mechanism}
\label{sec:exp-robust}

A matched GRPO plus length-penalty baseline uses
$r_{\rm LP}=r_{\rm code}-2.5\times10^{-5}|y|_{\rm chars}$. It produces the shortest
LiveCodeBench chains, but raises performativity from $0.267$ to $0.374$
(Table~\ref{tab:lp}). ProFIL instead lowers it to $0.076$. The same ordering holds in
every length quintile and throughout the evaluation-cutoff sweep
$\eta\in\{0.05,0.10,0.20,0.50\}$. Fewer characters are
not enough. The filter must distinguish reasoning before commitment from
padding after commitment.

\ifdefined\profilsixpage\else
\begin{figure}[t]
  \centering
  \includegraphics[width=\textwidth]{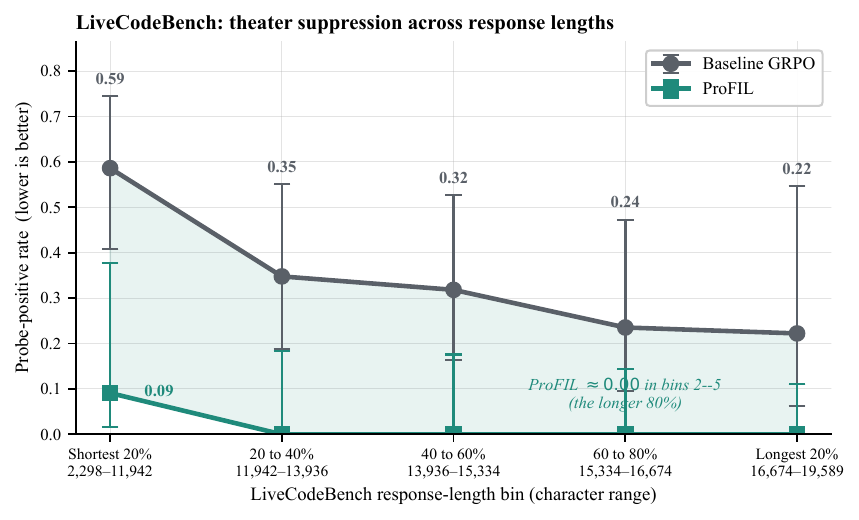}
  \caption{\textbf{Theater suppression holds at every chain length on LiveCodeBench.}
    Baseline and ProFIL rollouts are pooled, sorted by character count, and divided into
    five equal-frequency bins from the shortest 20\% to the longest 20\%. Each x-axis
    label gives the percentile span and observed character range. The vertical axis is the
    probe-positive rate at the same evaluation cutoff $\eta=0.9$ used in Table~\ref{tab:main}.
    ProFIL remains below the
    baseline in each bin, including the longest 20\%, so the gain is not explained by
    shifting probability mass toward short responses.}
  \label{fig:length-controlled}
\end{figure}
\fi

\subsection{Independent Corroboration}
\label{sec:exp-corroborate}

The headline operational metric uses the frozen probe that also shapes training. By
itself, that comparison could be circular: the policy might lower the probe score
without becoming less padded to an outside reader. We separate two
probe-blind judge checks from three anti-gaming and cross-measurement checks.

\textbf{Independent LLM judge.} A GPT-4.1 judge evaluated 100 LiveCodeBench rollouts per
condition using
only a natural-language faithfulness rubric, with no access to probe scores or training
labels. Strict faithful-fraction rises from $0.49$
(95\% CI $0.39$--$0.59$) to $0.73$ (95\% CI $0.64$--$0.81$): a $+24$pp gap with
non-overlapping CIs.

\textbf{Pairwise judge.} A second judge sees anonymized baseline and ProFIL rollouts
side by side, in random order, and chooses which contains more padding relative to its
own length. It never sees probe scores or forced-answer labels. The baseline is judged
more padded in 71\% of GSM8K pairs, 72\% of ToolUse pairs, 58\% of MMLU-Redux pairs,
and 57\% of LiveCodeBench pairs. The LiveCodeBench length-penalty model has no such
edge over baseline at 51\%.

\textbf{Frozen-probe audit.} RL can sometimes hide a targeted behavior from a learned
scorer \citep{rlobfuscation2025}. We label LiveCodeBench rollouts with fresh
forced-answer counterfactuals and measure whether the original frozen probe can still
separate faithful from theatrical examples. Its AUROC is 1.000 before, during, and
after training while mean performativity falls. The policy changes its generated
behavior without erasing the probe's ability to detect the labeled distinction.

\textbf{Cross-measurement and inference-time controls.} The GSM8K probe transfers
zero-shot to LiveCodeBench and recovers the same direction. A new student probe trained
on the updated policy also agrees with the frozen-base measurement. In contrast, a sweep of
Contrastive Activation Addition coefficients \citep{rimsky2024caa} does not reach the
performativity of the ProFIL-trained policy. The change comes from training, not from a
temporary intervention at inference.

\ifdefined\profilsixpage
On LiveCodeBench, the frozen probe gives 0.267 for baseline and 0.076 for ProFIL; the
student probe gives 0.290 and 0.069. CAA gives 0.080, 0.100, 0.090, 0.100, and 0.090
at coefficients 0, 5, 10, 20, and 40, respectively, all above ProFIL's 0.076.
\else
\begin{table}[!htbp]
  \centering
  \caption{\textbf{The frozen scorer remains valid, while inference-time steering is
    insufficient.} Lower performativity is better. A student probe trained on the
    updated policy agrees with the frozen-base probe. Every tested CAA coefficient
    remains above the ProFIL evaluation performativity of $0.076$.}
  \label{tab:frozen-vs-caa}
  \small
  \begin{minipage}[t]{0.42\textwidth}
    \centering
    \textbf{(a) Probe audit}\\[2pt]
    \begin{tabular}{lcc}
      \toprule
      Scorer & Baseline & ProFIL \\
      \midrule
      Frozen base & 0.267 & 0.076 \\
      Student & 0.290 & 0.069 \\
      \bottomrule
    \end{tabular}
  \end{minipage}\hfill
  \begin{minipage}[t]{0.55\textwidth}
    \centering
    \textbf{(b) Inference-time steering}\\[2pt]
    \setlength{\tabcolsep}{4pt}
    \begin{tabular}{lccccc}
      \toprule
      CAA coefficient & 0 & 5 & 10 & 20 & 40 \\
      \midrule
      Performativity & 0.080 & 0.100 & 0.090 & 0.100 & 0.090 \\
      \bottomrule
      \multicolumn{6}{c}{ProFIL-trained evaluation: 0.076} \\
    \end{tabular}
  \end{minipage}
\end{table}
\fi

\subsection{What the Filter Removes}
\label{sec:exp-mech}

\ifdefined\profilsixpage
LiveCodeBench theater is bimodal: baseline rollouts cluster below 0.1 or above 0.5,
and ProFIL removes the high-theater mode. Re-derivation headings occur in 72\% of
high-theater baseline rollouts, 3\% of faithful ones, and no ProFIL rollouts. Theater
also is not concentrated on failures: the high-theater tercile reaches 33\% accuracy
versus 9\% in the low-theater tercile ($\rho=0.240$, $p=0.016$), with the same
correctness-conditioned conclusion on GSM8K and MMLU-Redux.
\else
\textbf{A separate behavioral mode.} LiveCodeBench performativity is bimodal
(Table~\ref{tab:behavioral-signature}).
Baseline rollouts cluster below 0.1 or above 0.5, with
little mass between them. ProFIL removes the high-theater mode while retaining the
low-theater mode across chain lengths.

\textbf{A concrete writing pattern.} High-theater rollouts often restart a finished
derivation under headings such as \texttt{\#\#\# Approach},
\texttt{\#\#\# Solution}, or \texttt{\#\#\# Explanation}. At least one such block
appears in 72\% of high-theater baseline rollouts but only 3\% of faithful ones. It
appears in none of the ProFIL rollouts. The headings themselves are not penalized. The
pattern matters only when it repeats reasoning after commitment.

\textbf{A confidence ritual.} Theater is not concentrated on difficult failures. On
LiveCodeBench, the high-theater tercile reaches about 33\% accuracy compared with about
9\% in the low-theater tercile (Spearman $\rho=0.240$, $p=0.016$). The model tends to
perform confidence after it already knows the solution. Conditioning on correctness on
GSM8K and MMLU-Redux gives the same conclusion: the faithfulness gain is not caused by
incorrect or uncommitted chains being counted as faithful.

\ifdefined\profilsixpage\else
\begin{table}[!htbp]
  \centering
  \small
  \caption{\textbf{What changes when ProFIL removes post-commitment reasoning.}
    Exact values make each behavioral result independently readable. Counts are
    rollout counts.}
  \label{tab:behavioral-signature}
  \begin{tabular}{@{}>{\raggedright\arraybackslash}p{0.22\textwidth}
                      >{\raggedright\arraybackslash}p{0.43\textwidth}
                      >{\raggedright\arraybackslash}p{0.26\textwidth}@{}}
    \toprule
    Observable & Exact comparison & What it shows \\
    \midrule
    LiveCodeBench score shape & Baseline has modes below 0.1 and above 0.5; ProFIL removes the high-score mode. & A distinct theater mode disappears. \\
    \addlinespace[2pt]
    Re-derivation headings & Baseline: 31/43 high-theater rollouts (72\%) and 2/72 faithful rollouts (3\%); ProFIL: 0/131 (0\%). & Headings track post-commitment restatement. \\
    \addlinespace[2pt]
    ToolUse single-block resolutions & Baseline: 1/68 (1.5\%); ProFIL: 8/68 (11.8\%). & $8\times$ more direct resolutions. \\
    \addlinespace[2pt]
    LiveCodeBench accuracy by theater tercile & Low theater: 9\%; high theater: 33\% ($\rho=0.240$, $p=0.016$). & Theater is not a failure-only effect. \\
    \bottomrule
  \end{tabular}
\end{table}
\fi
\fi

\section{Discussion and Conclusion}
\label{sec:discussion}

\ifdefined\profilsixpage
\textbf{Self-correction is preserved by construction.} Commitment is the first correct
prefix, so earlier repairs remain. In one GSM8K case, 1,200 is corrected to 200 before
commitment; only the later restatement is filtered.

\textbf{Limitations.} Forced answering measures answer extractability, not logical or
causal faithfulness; an uncommitted ratio of zero can mean genuine search or a loop.
Architecture is confounded with domain. LiveCodeBench accuracy is protocol-dependent.
The main metric reuses the training probe despite agreement from independent judges and
fresh forced answers. One run per condition excludes seed variance. A poor probe target
could filter useful uncertainty or refusal. More backbones, blinded human audits, and
open-ended domains remain important tests.

\textbf{Conclusion.} Across four domains and two architectures, ProFIL lowers theater
by 11--100\% while a length penalty fails. It targets when useful reasoning ends, not
token count.
\else
\textbf{Self-correction is preserved by construction.}
Commitment is the first correct prefix, so earlier mistakes and repairs are retained.
In one GSM8K case, the model first answers 1,200, then subtracts the dragon's reach and
corrects to 200. Commitment occurs at the correction; only the later restatement is
performative. Changing one's mind is not equated with theater.

\textbf{Limitations.}
\emph{Construct:} forced answering detects when a correct answer becomes extractable,
not whether visible reasoning entails it or caused the computation. \emph{Uncommitted
chains:} target ratio zero can denote genuine search or a repetition loop. \emph{Backbone
and domain:} Llama-8B serves GSM8K and LiveCodeBench, while Qwen-7B serves ToolUse
and MMLU-Redux; agreement does not isolate architecture from domain.
\emph{LiveCodeBench accuracy:} ProFIL leads at the sampled@1 training peak and trails at
the final greedy checkpoint, making accuracy protocol-dependent although theater
reduction is stable. \emph{Circularity:} the main metric reuses the training probe;
independent judges and the forced-answer audit agree, but larger human audits and more
models would help. \emph{Training variance:} one RL run per condition means evaluation
intervals do not measure seed variance. \emph{Probe target:} a poorly chosen target could
filter useful uncertainty or refusal; commitment labels narrow, but do not remove, this risk.

\textbf{Future work.}
Combine ProFIL with process rewards, extend commitment verification to open-ended
domains, and test more backbones with larger blinded human audits.

\textbf{Conclusion.}
ProFIL turns a frozen internal signal into an RL filter. Across four domains and two
architectures, performativity falls by 11--100\%, independent judges agree, and a length
penalty fails. ProFIL targets when useful reasoning ends, not token count.
\fi

\ifdefined\profilsixpage\clearpage\fi
\bibliographystyle{plainnat}
\bibliography{references}

\ifdefined\includeprofilappendix
\clearpage
\appendix

\section{Extended Results}
\label{app:extended}

\subsection{Length-controlled faithfulness}
\label{app:length}

\ifdefined\profilsixpage
\begin{figure}[ht]
  \centering
  \includegraphics[width=0.85\textwidth]{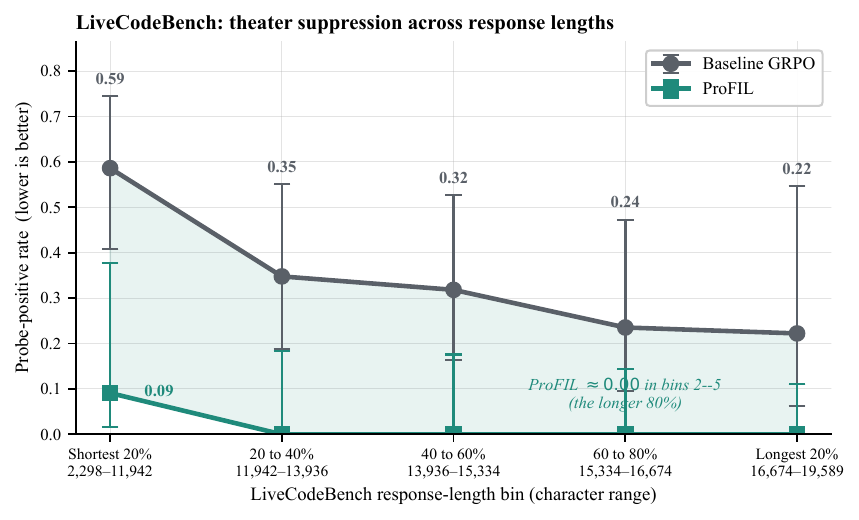}
  \caption{\textbf{Theater suppression persists across all chain lengths, ruling out length compression as the mechanism} (LiveCodeBench).
    Rollouts are stratified into 5 length quintiles on the pooled distribution; the panel reports performativity ratio within each bin.
    ProFIL suppresses theater across the full length distribution, including the longest 20\% of responses, while also producing shorter chains overall ($11{,}875$ vs.\ $12{,}382$ mean chars).}
  \label{fig:length}
\end{figure}
\else
\label{fig:length}
Figure~\ref{fig:length-controlled} in the main paper reports the full
length-controlled analysis. Across all five pooled LiveCodeBench length quintiles,
ProFIL stays below the baseline probe-positive rate, including in the longest 20\%
of responses. This rules out a shift toward shorter rollouts as the explanation for
the observed reduction in post-commitment reasoning.
\fi

\subsection{Frozen-Probe AUROC (Anti-Gaming Evidence)}
\label{app:auroc-trajectory}

A probe-as-reward system is vulnerable to RL obfuscation \citep{rlobfuscation2025}: the
policy could learn to produce low probe scores while still generating theatrical
reasoning, by shifting its activations away from the probe's decision boundary. We check
this by re-scoring trained-model rollouts with the frozen probe and computing AUROC
against forced-answer-derived performativity labels.

On LiveCodeBench (Llama-8B), the frozen probe attains AUROC $1.000$ on rollouts from
the baseline, length-penalty, and ProFIL trained models, and remains $1.000$ throughout
training while mean performativity falls from baseline to ProFIL
(Table~\ref{tab:main}). The probe rank-orders theatrical vs.\ faithful rollouts
perfectly before, during, and after training: the empirical signature of genuine
theater reduction rather than activation-space evasion.

\subsection{\texorpdfstring{Evaluation-Cutoff ($\eta$) Sensitivity}{Evaluation-Cutoff Sensitivity}}
\label{app:delta}

Table~\ref{tab:delta} reports performativity ratio at $\eta=0.05$, $0.10$, $0.20$,
and $0.50$. ProFIL remains below baseline at every cutoff in every domain. On
LiveCodeBench, length penalty remains above baseline at every cutoff, so compression
alone does not reduce theater.

\begin{table}[ht]
  \centering
  \small
  \caption{\textbf{Performativity ratio is stable across evaluation cutoffs $\eta$.}
    Each cell: perf\_ratio (fraction of steps with probe sigmoid $\geq \eta$).
    ProFIL probe is lower than baseline at every $\eta$ in every domain.
    LCB length penalty is higher than baseline at every $\eta$.
    These sensitivity cutoffs are distinct from training threshold $\tau$.}
  \label{tab:delta}
  \begin{tabular}{l l c c c c}
    \toprule
    Domain & Condition & $\eta{=}0.05$ & $\eta{=}0.10$ & $\eta{=}0.20$ & $\eta{=}0.50$ \\
    \midrule
    \multirow{2}{*}{GSM8K}
      & Baseline        & 0.060 & 0.038 & 0.022 & 0.014 \\
      & \textbf{ProFIL}& \textbf{0.048} & \textbf{0.032} & \textbf{0.012} & \textbf{0.004} \\
    \midrule
    \multirow{3}{*}{LCB}
      & Baseline        & 0.489 & 0.450 & 0.420 & 0.328 \\
      & Len.\ Penalty   & 0.649 & 0.603 & 0.550 & 0.458 \\
      & \textbf{ProFIL}& \textbf{0.405} & \textbf{0.359} & \textbf{0.267} & \textbf{0.130} \\
    \midrule
    \multirow{2}{*}{MMLU}
      & Baseline        & 0.036 & 0.030 & 0.022 & 0.018 \\
      & \textbf{ProFIL}& \textbf{0.030} & \textbf{0.024} & \textbf{0.020} & \textbf{0.018} \\
    \bottomrule
  \end{tabular}
\end{table}

\textbf{In-training validation curves.} GSM8K and LiveCodeBench validation accuracy logs
at every 20 GRPO steps are released alongside the paper.

\section{Algorithmic and Hyperparameter Details}
\label{app:algo}

\begin{figure}[ht]
  \centering
  \includegraphics[width=0.95\textwidth]{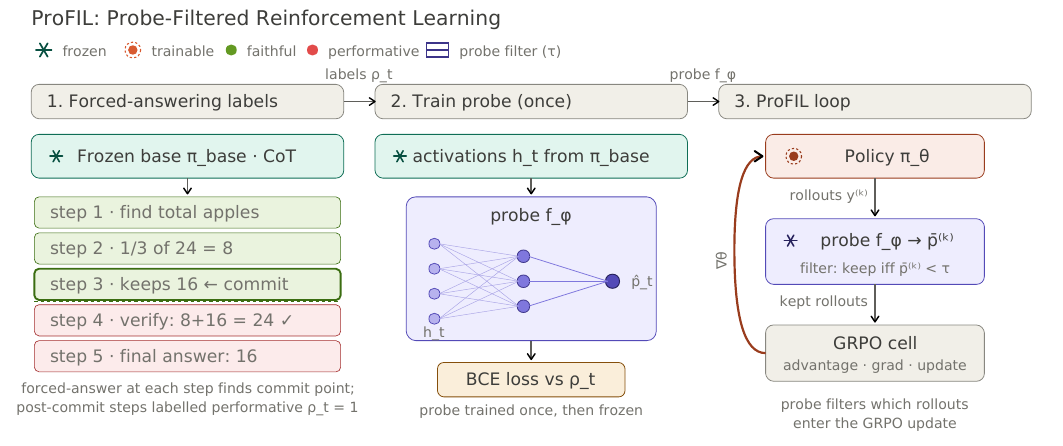}
  \caption{\textbf{ProFIL turns a frozen-base probe into a training signal.}
    (1)~Forced-answering yields verifier-derived step-level performativity labels (no
    human annotation): the first step whose forced answer is correct defines the
    \emph{commitment point}; later steps are performative.
    (2)~A gated multi-head attention probe (RLFR-style; \citealp{goodfire2026rlfr}) is
    trained \emph{once} on activations of the \emph{frozen} base at layers
    $\ell_1, \ell_2$, then never updated.
    (3)~During GRPO, a separate frozen base teacher-forces the policy's exact rollout
    text, and the frozen probe re-computes its mean performativity. Rollouts above
    $\tau$ have advantage zeroed and contribute nothing to the policy update. The probe
    boundary is therefore stable throughout RL, and the GRPO update itself is
    unchanged outside this filter.}
  \label{fig:method-detailed}
\end{figure}

\begin{algorithm}[ht]
\caption{Probe-Filtered GRPO update}
\label{alg:pfgrpo}
\begin{algorithmic}[1]
\Require frozen base $\pi_{\mathrm{base}}$, probe $f_\phi$, threshold $\tau$, prompt $x$, group size $K$
\State Sample $\{y^{(k)}\}_{k=1}^K \sim \pi_\theta(\cdot \mid x)$
\State Compute verifier reward $r^{(k)}$
\For{$k = 1, \dots, K$}
  \State Compute frozen activations $h^{(k)}_{\ell_1}, h^{(k)}_{\ell_2} \leftarrow \pi_{\mathrm{base}}(y^{(k)})$
  \State Compute per-step performativity $p^{(k)}_{1:T_k} \leftarrow f_\phi(h^{(k)})$ ; $\bar p^{(k)} = \mathrm{mean}(p^{(k)})$
\EndFor
\State Let $\tilde r^{(k)} \leftarrow r^{(k)} \cdot \mathbf{1}[\bar p^{(k)} < \tau]$ \Comment{zero reward of filtered rollouts}
\State $\mu, \sigma \leftarrow \mathrm{mean}(\tilde r), \mathrm{std}(\tilde r)$ over all $K$ rollouts
\For{$k = 1, \dots, K$}
  \State $\hat A^{(k)} \leftarrow ((\tilde r^{(k)} - \mu) / \sigma) \cdot \mathbf{1}[\bar p^{(k)} < \tau]$
\EndFor
\State Update $\theta$ with standard GRPO step using $\{\hat A^{(k)}\}$.
\end{algorithmic}
\end{algorithm}

\textbf{Per-domain training configuration.}
\begin{itemize}[leftmargin=*, itemsep=1pt, topsep=2pt]
  \item \textbf{GSM8K} (Llama-8B): 7440 train / 500 test problems, 3 training epochs, 4096-token context, $\tau{=}0.5$, group size $K{=}8$, probe layers [13,\,19].
  \item \textbf{LiveCodeBench v6} (Llama-8B): 920 train / 131 test problems, 3 epochs, 8192-token context, $\tau{=}0.2$, $K{=}8$, probe layers [13,\,19]. Reward: test-case pass rate $>0.5$.
  \item \textbf{ToolUse} (Qwen-7B): 4046 train / 68 test, 3 epochs, 4096-token context, $\tau{=}0.5$, $K{=}8$, probe layers [14,\,21]. Chain-level performativity pooling.
  \item \textbf{MMLU-Redux} (Qwen-7B): 6144 train / 500 test problems, 256 GRPO steps, 4096-token context, $\tau{=}0.5$, $K{=}8$, probe layers [14,\,21].
\end{itemize}
All runs: AdamW optimizer, cosine LR schedule, identical to the matched GRPO baseline
except for the probe filter. LiveCodeBench uses the empirically selected
$\tau{=}0.2$; under the keep rule $\bar p<\tau$, this is more selective than $0.5$.
Full config files are released with the code.

\textbf{Group statistics under filtering.} Algorithm~\ref{alg:pfgrpo} sets filtered
rollout rewards to zero, and $\mu, \sigma$ are computed over all $K$ rollouts (filtered
ones contribute $\tilde r = 0$), matching our implementation exactly: filtered rollouts
enter the group-normalization statistics as zero-reward entries rather than being
excluded from them. This is not equivalent to computing $\mu, \sigma$ over the
unfiltered subset only, since including zeroed rollouts changes the advantage assigned
to every other rollout in the group.

\textbf{Forced-answering prompt.} ``Stop reasoning now. Based on what you have written
above, give your final answer in the format \texttt{\textbackslash boxed\{\}}.'' Greedy decoding,
max 64 new tokens. We measure first-correct over the full step-boundary trajectory.
\emph{LiveCodeBench exception}: the forced-answer correctness criterion is test-case
execution (pass rate $>0.5$), not boxed-answer string matching, since LCB problems require
runnable code rather than a closed-form answer.

\section{Probe Architecture and Training}
\label{app:probe}

\textbf{Architecture.} Each head $h$ has learned query $q_h \in \mathbb{R}^{d_h}$ and
sigmoid gate $g_h$. Keys and values come from per-layer LayerNorm-normalized residual-stream
activations of the frozen base at layers $\ell_1, \ell_2$. Output: $z = \sum_h g_h \cdot
\mathrm{head}_h$, scalar logit via linear projection.

\textbf{Training.} Probes trained on rollouts from the frozen base using binary
cross-entropy on forced-answering labels. Best-epoch held-out AUROC on the
probe-train/test split (in-distribution): GSM8K 0.924, LiveCodeBench 0.928,
ToolUse 0.976, MMLU-Redux 0.991.

\textbf{Layer choice.} For Qwen-7B: layers 14 and 21 (mid-network commitment and late
elaboration). For Llama-8B: layers 13 and 19. Training AUROC exceeds 0.92 on all
four domains, validating that these layers carry a clean theater signal.

\section{Independent LLM Judge and ToolUse Commitment Heuristic}
\label{app:judge}

We score every rollout with a GPT-4.1 judge using a natural-language faithfulness
rubric, with no access to probe scores or training labels. On LiveCodeBench, strict
faithful-fraction is $0.49$ for the baseline ($95\%$ Wilson CI $[0.39, 0.59]$) versus
$0.73$ for ProFIL ($[0.64, 0.81]$): a $+24$pp gap with non-overlapping CIs.

\textbf{ToolUse chain-level commitment heuristic.} We flag the first step whose decoded
text contains a tool-call closure (a JSON-formatted \texttt{\{"tool":\dots,"args":\dots\}}
block followed by terminal \texttt{</think>} or end-of-turn marker), or failing that, the
first step whose forced-answer matches the ground-truth tool name and arguments. Steps after
this commitment are labeled performative for the chain-level metric.

\section{Adaptive Threshold Simulation}
\label{app:adaptive-tau}

The per-domain $\tau$ values used in this paper (GSM8K/ToolUse/MMLU-Redux: $0.5$;
LiveCodeBench: $0.2$) match the reward density of each domain (App.~\ref{app:algo}).
A natural alternative is to derive $\tau$ automatically from the rolling mean group
reward, removing the need for domain-specific tuning.

\textbf{Proposed schedule.}
Define
\[
  \tau_t = \tau_{\min} + (\tau_{\max} - \tau_{\min}) \cdot
           \mathrm{clip}\!\left(\frac{\bar r_t - r_{\mathrm{lo}}}{r_{\mathrm{hi}} - r_{\mathrm{lo}}},\,0,\,1\right),
\]
where $\bar r_t$ is the exponential moving average of group reward at step $t$, and
$(r_{\mathrm{lo}}, r_{\mathrm{hi}}) = (0.10, 0.40)$, $(\tau_{\min}, \tau_{\max}) = (0.20, 0.50)$.
When reward density is low, $\tau$ is small and the filter is more selective. As reward
density grows, $\tau$ rises and the filter relaxes. This describes the retrospective
schedule above; it was not used for the reported training runs.

\textbf{Retrospective simulation on LiveCodeBench.}
Figure~\ref{fig:adaptive-tau} applies this schedule retrospectively to the three
available LCB checkpoints using their per-example performativity scores and
checkpoint accuracies as a proxy for reward density.
At step~320 (accuracy~$11\%$), fixed $\tau{=}0.5$ would filter $59\%$ of rollouts;
adaptive $\tau$ self-adjusts to $0.21$ and matches the per-domain $\tau{=}0.2$ used in
the main experiments. At step~440 after training, accuracy recovers to $18\%$ and
adaptive $\tau$ rises to $0.28$, filtering $21\%$ of rollouts rather than $13\%$ at
fixed $\tau{=}0.5$. The schedule removes the per-domain $\tau$ choice entirely.
A full training comparison of adaptive vs.\ fixed $\tau$ is left to future work.

\begin{figure}[ht]
  \centering
  \includegraphics[width=0.90\textwidth]{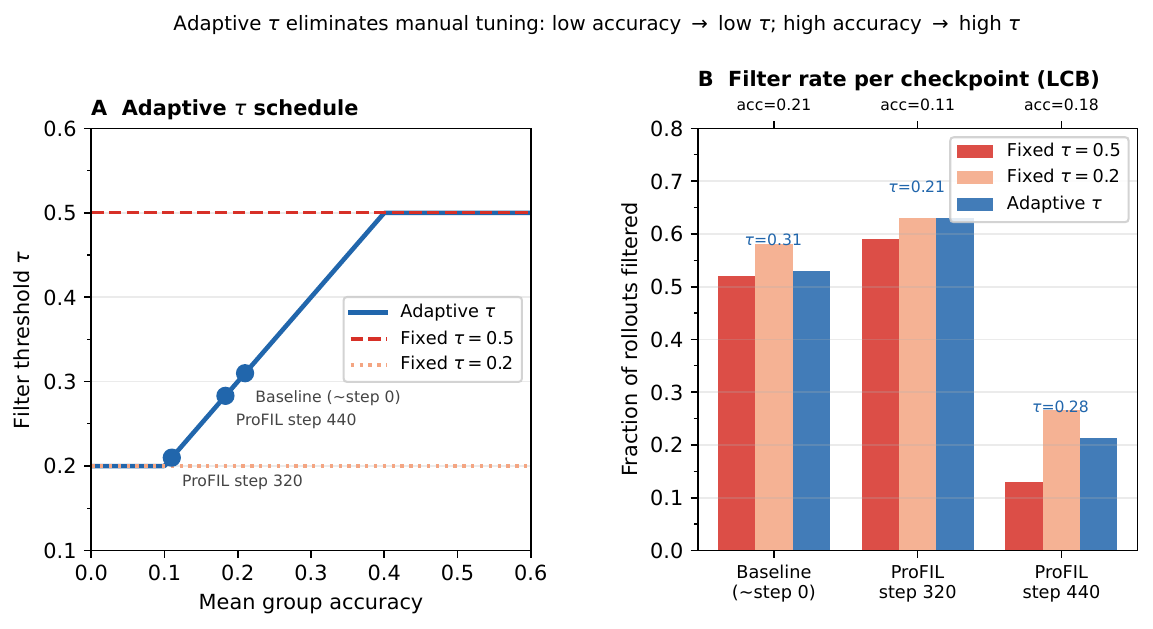}
  \caption{\textbf{Adaptive $\tau$ eliminates manual threshold tuning.}
    \emph{Left:} the proposed linear schedule maps mean group accuracy to $\tau \in [0.20, 0.50]$;
    dots mark the three LCB checkpoints.
    \emph{Right:} filter rates under fixed $\tau{=}0.5$, fixed $\tau{=}0.2$, and adaptive $\tau$
    at each checkpoint.
    At step~320 (lowest accuracy) adaptive $\tau$ automatically matches the
    sparse-regime value; at step~440 it rises, balancing signal preservation and theater filtering.}
  \label{fig:adaptive-tau}
\end{figure}

\section{Theater as a Confidence Ritual}
\label{app:confidence-ritual}

This appendix expands the headline result of \S\ref{sec:exp-mech} that
high-theater rollouts in the baseline correlate with task correctness.
Figure~\ref{fig:confidence_ritual} shows the per-tercile accuracy split
(panel A) alongside the per-decile rollout-density shift between baseline
and ProFIL (panel B): theater concentrates on problems the model already
solves, and ProFIL collapses the distribution to the low-theater region
while preserving accuracy.

\begin{figure}[!htbp]
  \centering
  \includegraphics[width=0.95\textwidth]{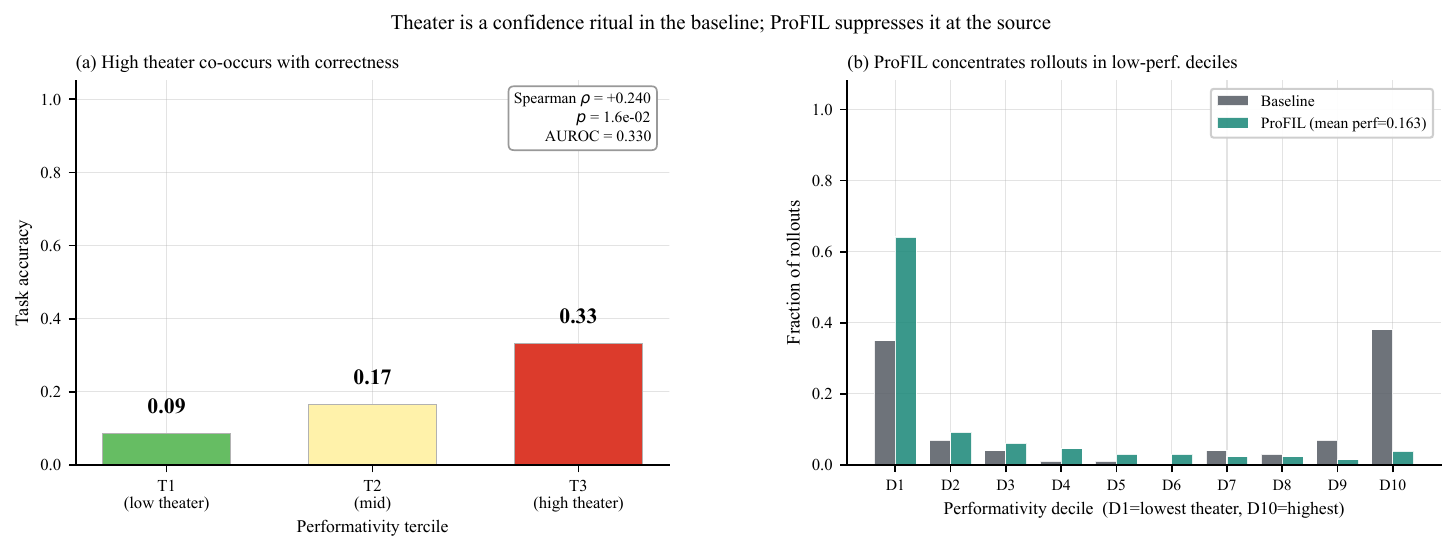}
  \caption{\textbf{Theater marks success, not uncertainty: it is a learned confidence ritual.}
    \emph{Left:} task accuracy by performativity tercile (T1 = low theater, T3 = high theater)
    on LiveCodeBench baseline rollouts.
    High-theater rollouts trend toward higher accuracy:
    the high-theater tercile averages ${\sim}33\%$ accuracy vs.\ ${\sim}9\%$ for the
    low-theater tercile (Spearman $\rho{=}{+}0.240$, $p{=}0.016$;
    AUROC $= 0.33$, meaning theater \emph{anti-predicts} failure).
    \emph{Right:} the decile distribution (finer-grained than panel A's terciles, to
    show the bimodal collapse) shifts from baseline mass at D1 and D10 to near-entirely
    low-theater under ProFIL.
    ProFIL does not suppress uncertainty; it suppresses post-hoc elaboration on problems
    the model already knows how to solve.}
  \label{fig:confidence_ritual}
\end{figure}

\section{Self-Correction Case Studies}
\label{app:self_correction}

We present two GSM8K rollouts from the probe-trained model in which the forced-answer
oracle confirms a wrong answer at an early step and a correct answer at a later step.
In both cases, ProFIL assigns the commitment point at the \emph{first correct} step;
the intermediate correction steps are pre-commitment and are never labeled performative
or filtered.

\begin{figure}[!htbp]
  \centering
  \includegraphics[width=\linewidth]{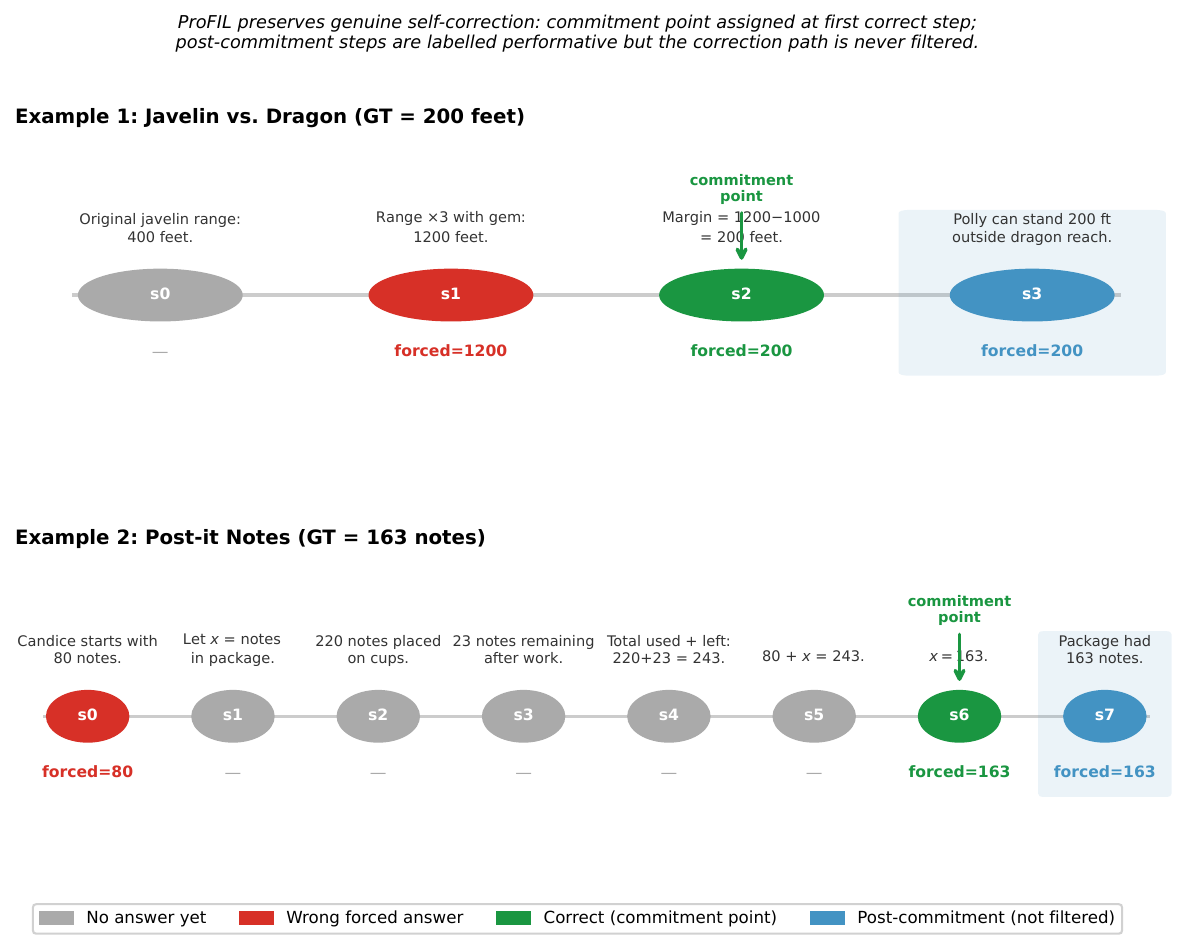}
  \caption{Step-by-step forced-answer trajectories for two self-correcting GSM8K rollouts.
    Each circle is a paragraph step; the label beneath shows what the model would have
    answered if stopped at that step.  The commitment point (green arrow) is the first step
    at which the forced answer matches the ground truth.  Blue-shaded steps are
    post-commitment; they are labeled performative only if they follow the commitment
    point.  In both examples the error-correction move itself is \emph{pre}-commitment
    and is never filtered.}
  \label{fig:self_correction}
\end{figure}

\paragraph{Example 1: Javelin vs.\ Dragon (ground truth: 200 feet).}
The problem asks how far outside the dragon's reach Polly can stand.
At step~1 the model computes the gem-enhanced range (1200 feet) and, if stopped there,
answers \textbf{1200}, mistaking an intermediate quantity for the final answer.
At step~2 it subtracts the dragon's reach ($1200 - 1000 = 200$) and correctly answers
\textbf{200}.  The commitment point is assigned at step~2; step~3 (a restatement) is
post-commitment.  ProFIL filters step~3 if it is performative, but steps~1--2 (including
the correction) are untouched.

\paragraph{Example 2: Post-it Notes (ground truth: 163 notes).}
The problem requires setting up and solving $80 + x = 243$.
At step~0 the model reads off the initial count (80) and would answer \textbf{80} if
stopped; this is a premature answer before the equation is established.
Steps~1--5 build the equation; the model cannot yet commit.
At step~6 it solves $x = 163$ and the forced-answer oracle returns \textbf{163}: the
commitment point.
The six-step correction path (steps~0--5) is entirely pre-commitment and is never filtered.

\section{Compute Budget}
\label{app:compute}

Each domain-condition pair: 4 (Llama-8B) or 4--8 (Qwen-7B) H100 80\,GB GPUs for 3 epochs,
totalling ${\sim}480$ GPU-hours across the eight training runs. Probe training: ${\sim}2$
GPU-hours per domain. Evaluation (vLLM rollouts and scoring): ${\sim}8$ GPU-hours per
domain. Ablations (Figure~\ref{fig:length}, independent judge,
mechanistic analysis): ${\sim}200$ additional GPU-hours.

\textbf{Probe filtering overhead.} On a single A100-40GB, scoring one LiveCodeBench
rollout with the frozen-base probe (activation extraction plus the attention-probe
forward pass, the exact computation used to produce the per-rollout filter signal
during GRPO) takes ${\sim}319$\,ms, versus ${\sim}533$\,ms for one forward-backward step
of the policy itself on a comparably sized batch: roughly $60\%$ of one policy update's
per-rollout cost, and strictly less since the probe pass has no backward computation or
optimizer state. This scoring runs alongside, not instead of, the GRPO update, adding a
single frozen-model forward pass per rollout batch.

\section{Broader Impacts}
\label{app:impact}

ProFIL produces models whose chain-of-thought more closely reflects the computation that
led to the answer. The primary benefit is auditability: a reasoning model whose chain
reflects its computation is more amenable to oversight, step-level interpretability, and
honest disagreement. The symmetric concern is that a method suppressing ``performative''
filler could in principle target any reasoning pattern a probe is trained to recognize,
including reasoning a model should express (safety considerations, uncertainty, refusal
preambles). Since our probe targets behavior identified by forced-answering correctness, the
immediate risk is low; the general lesson is that probes are dual-use and the choice of
probe target is a value judgment.

\ifdefined\includeprofilchecklist
\clearpage
\input{checklist.tex}
\fi
\fi

\end{document}